\documentclass[runningheads]{llncs}

\usepackage{eccv}

\usepackage{eccvabbrv}

\usepackage{graphicx}
\usepackage{booktabs}

\usepackage[accsupp]{axessibility}  

\usepackage[pagebackref,breaklinks,colorlinks,citecolor=eccvblue]{hyperref}

\usepackage{orcidlink}

\usepackage{booktabs}

\usepackage{algorithm}
\usepackage{algpseudocode}

\usepackage{multirow}
\usepackage[table]{xcolor}
\usepackage{pifont}
\usepackage{enumitem}
\usepackage{soul}
\usepackage{tcolorbox}

\newcommand{\cmark}{\ding{51}}

\begin{document}

\title{
Diagnosing Aerial-View Object Detectors with Foundational Image Generative Models
}

\titlerunning{Diagnosing Aerial-View Detectors using Generative Models}

\author{
Stanislav Panev\inst{1}\orcidlink{0000-0002-2193-5847} \and
Minhyek Jeon\inst{1}\orcidlink{0000-0001-5208-5921} \and
Vaishnavi Khindkar\inst{1}\orcidlink{0000-0001-6058-9089} \and \\
Ahish Deshpande\inst{1} \and
Celso M de Melo\inst{2} \and
Shuowen Hu\inst{2} \and \\
Shayok Chakraborty\inst{1,3}\orcidlink{0000-0001-6378-8286} \and 
Fernando De la Torre\inst{1}
}

\authorrunning{S.~Panev \etal}

\institute{
    Carnegie Mellon University, USA \\
    \email{spanev@andrew.cmu.edu},
    \email{minhyekj@alumni.cmu.edu}, \\
    \email{\{vkhindka, ahishd\}@andrew.cmu.edu}, \email{ftorre@cs.cmu.edu} \and
    DEVCOM Army Research Laboratory, USA\\
    \email{\{celso.m.demelo.civ, shuowen.hu.civ\}@army.mil} \and
    Florida State University, USA
    \email{shayok@cs.fsu.edu}
}

\maketitle

\begin{abstract}

Recent advances in large-scale image generative models enable photorealistic scene synthesis with controllable attributes. Beyond data augmentation, their potential as diagnostic tools for trained vision systems remains unexplored in the aerial and remote sensing domains.
We introduce a synthetic diagnostic framework for aerial-view vehicle detection that combines text-guided generation, attribute-controlled editing, and automated attribute verification to construct a controllable synthetic testbed. 
This enables fine-grained evaluation of pretrained detectors under diverse scene types and environmental conditions that are difficult to isolate in real datasets.
Across three detection architectures and three real aerial datasets, synthetic scene-wise performance trends closely match real-world weaknesses. 
Guided by these diagnostics, targeted supplementation with small real datasets from the identified weak categories yields improvements of up to 13\% AP50 while requiring substantially fewer additional samples than non-targeted augmentation.
Our results show that controlled synthetic probing can predict real-domain performance gaps and guide efficient data collection. The proposed diagnostic framework is modular and can incorporate alternative generative or vision-language models as capabilities evolve. 
Our code and datasets are available here: 
\href{https://humansensinglab.github.io/AVODDiag/}{humansensinglab.github.io/AVODDiag/}
\end{abstract}

\begin{figure}[t]
    \centering
    \includegraphics[width=\textwidth]{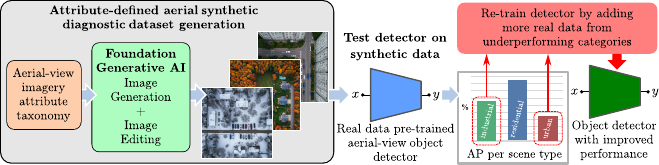}
    \caption{
      We leverage foundation generative models to synthesize diverse, attribute-controlled aerial-view image datasets, thereby enabling systematic diagnosis of the weaknesses of aerial-view vehicle detectors.
    }
    \label{fig:teaser}
\end{figure}
\section{Introduction}
\label{sec:intro}

Recent advances in foundation image generative models---such as \textit{Imagen~3}~\cite{baldridge2024imagen}, \textit{Gemini 2.5 Flash Image}~\cite{comanici2025gemini}, \textit{GPT-Image-1}~\cite{openai_gpt_image_1}, \textit{DALL$\cdot$E~3}~\cite{openai_dalle3}, and \textit{Midjourney}~\cite{midjourney}---enable the synthesis of highly realistic imagery with controllable semantic attributes. While these models are widely used for data augmentation, their potential as diagnostic tools for analyzing trained vision systems remains largely unexplored in remote sensing and aerial imagery. \textit{Aerial-view object detection} (AVOD) is a core component of \textit{Earth observation} (EO) pipelines that support disaster response, infrastructure inspection, defense intelligence, urban planning, and environmental monitoring. In such high-stakes applications, systematic performance degradation under specific scene conditions can directly affect downstream decision-making, motivating reliable and interpretable diagnostic evaluation.

Modern aerial benchmarks such as \textit{DOTAv2}~\cite{DOTAv2}, \textit{LINZ}~\cite{fang2025adapting}, and \textit{UGRC}~\cite{fang2025adapting} are inherently imbalanced: certain scene types and environmental conditions are overrepresented, while others are scarce or entangled with confounding factors. As a result, detectors may exhibit systematic performance degradation in specific environments (\eg, dense urban or industrial areas) that is difficult to isolate using standard aggregate metrics. Collecting geographically and environmentally balanced aerial datasets is costly and often infeasible, making controlled diagnostic evaluation a persistent challenge in remote sensing.

\begin{figure}[t]
    \centering
    \includegraphics[width=\linewidth]{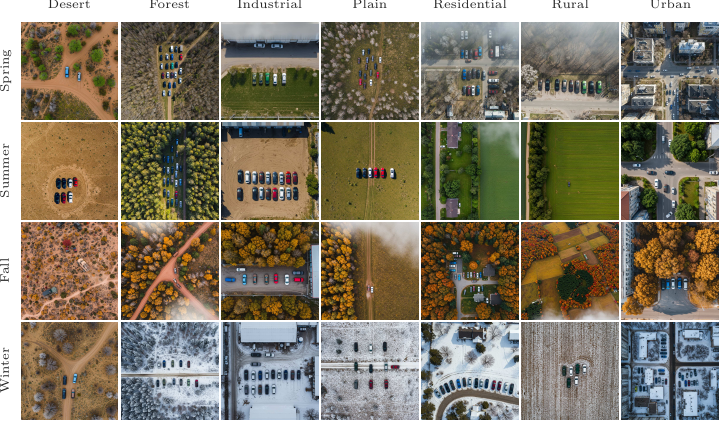}
    \caption{
        A showcase of synthetic aerial view imagery generated by \textit{Imagen 3}. 
        Each image represents a combination of a different environment and a season. We use these data to identify inherent weaknesses in object detectors.
    }
    \label{fig:synthetic_dataset}
\end{figure}

In this work, we investigate whether foundation image generative models can be leveraged to construct a controlled synthetic diagnostic testbed for aerial-view vehicle detection (\cref{fig:teaser}). 
As an illustrative example, \cref{fig:synthetic_dataset} shows synthetic images generated by \textit{Imagen 3} under different attribute settings.
Rather than proposing a new detector or generative model, we introduce a synthetic diagnostic framework that combines:
(1) text-guided aerial image generation,
(2) attribute-controlled image editing to reduce distributional imbalance, and
(3) automated attribute extraction and validation using multimodal models.
This process produces a fully synthetic dataset annotated with explicit environmental and object-level attributes, enabling scene-conditioned evaluation of pretrained detectors under controlled variations of scene type, season, weather, and object properties.
We evaluate three widely adopted detection architectures---
\textit{Faster R-CNN}~\cite{ren2015faster}, \textit{YOLOv8}~\cite{mmyolo2022}, and \textit{ViTDet}~\cite{li2022exploring}---trained on three real aerial datasets. Synthetic scene-wise performance trends consistently align with real-world weaknesses, with urban and industrial environments emerging as recurrent failure modes across models and training sources. Guided by these diagnostics, we perform targeted real-data supplementation by adding scene-specific images corresponding to the identified weak categories. 
Despite being at least an order of magnitude smaller than the original training sets, this targeted supplementation yields improvements of up to $13\%$ in AP50. 

Our contribution is therefore a validated synthetic-to-real diagnostic framework for aerial-view detection systems. We demonstrate that controlled synthetic probing can predict real-domain performance gaps and provide actionable guidance for efficient data acquisition in remote sensing applications. While our empirical study focuses on RGB aerial imagery and vehicle detection, the proposed diagnostic framework is modular and generalizable, and can integrate alternative generative and multimodal models as capabilities evolve, and can be applied to non-aerial imagery systems as well. 
In summary, our contributions are:
\begin{enumerate}
    \item A synthetic, attribute-controlled diagnostic framework for analyzing aerial-view object detectors using foundation generative models.
    \item Empirical evidence that synthetic scene-wise evaluation reliably reflects real-world performance weaknesses across multiple detectors and datasets.
	\item A targeted data supplementation strategy guided by synthetic diagnostics, achieving up to $13\%$ AP50 improvement with limited additional real data.
\end{enumerate}

\section{Related Work}
\label{sec:related_work}
\subsection{Model Diagnosis in Object Detection}

Traditional model diagnostics rely on confusion matrices and performance metrics, which are insufficient for characterizing systemic failure modes. Recent work moves beyond coarse summaries toward fine-grained failure analysis. Hoiem \etal~\cite{hoiem2012diagnosing} categorized detection errors into localization mistakes, confusions with similar classes, and background false positives. Building on this, Bolya~\etal~\cite{bolya2020tide} proposed \textit{TIDE}, decomposing detection errors into six types for fine-grained diagnosis, exposing previously hidden failure patterns. 

However, these diagnostic approaches remain post-hoc, where multiple environmental factors are entangled and difficult to control. Explainability methods such as attribution maps~\cite{Selvaraju_2019} and local surrogate explanations~\cite{ribeiro2016whyitrustyou} highlight decision-relevant regions but do not isolate the effect of scene-level attributes on detection performance. Synthetic datasets such as \textit{InteriorNet}~\cite{InteriorNet18} provide large-scale rendered environments yet lack sufficient attribute-level controllability, while corruption benchmarks~\cite{DBLP:journals/corr/abs-1903-12261} reveal sensitivity patterns under distribution shift without enabling causal attribute manipulation of scene conditions. In contrast, our method leverages controllable synthetic generation to isolate individual scene attributes while keeping others fixed, enabling causal and interpretable failure analysis infeasible with real-world data.

\subsection{Synthetic Data for Benchmarking}

Using synthetic imagery to evaluate and improve vision models has a long history. Early work used game engines for semantic segmentation~\cite{richter2016playing, 7780721} and object tracking~\cite{gaidon2016virtual}, but suffered domain gaps due to texture artifacts and unrealistic lighting. M\"{u}ller~\etal~\cite{muller2018sim4cv} improved photorealism with a physics-based \textit{Unreal Engine} simulator, yet its fixed scene configurations remain impractical for large-scale, attribute-controlled diagnostic generation. In the \textit{autonomous driving} domain, generative models have recently been used to synthesize realistic driving scenarios for evaluation and failure mode discovery~\cite{hu2023gaia1generativeworldmodel, nvidia2025cosmosworldfoundationmodel}. However, they are tailored to the temporal, multi-modal structure of driving data (video, LiDAR, RADAR) and do not transfer directly to aerial-view detection, which operates on single frames with unique viewpoint and scale characteristics.

Diffusion models have enabled scalable and photorealistic synthesis. For instance, \textit{DatasetDM}~\cite{wu2023datasetdm} generates labeled datasets but targets augmentation rather than diagnosis, while zero-shot evaluation approaches~\cite{li2023your, clark2023text} using \textit{Stable Diffusion}~\cite{rombach2022high} lack fine-grained attribute control. Methods combining diffusion with \textit{NeRF}s~\cite{Poole2022DreamFusionTU} or \textit{3D Gaussian Splatting}~\cite{kerbl20233d} achieve high-fidelity, 3D-consistent synthesis; however, because they represent appearance as an entangled function of 3D geometry, lighting, and texture, they do not support the manipulation of individual 2D appearance attributes that is necessary for controlled diagnostic analysis. More recently, large multimodal models with integrated generation such as \textit{ChatGPT} with \textit{DALL$\cdot$E~3}~\cite{betker2023improving} and \textit{Gemini}'s \textit{Imagen~3}~\cite{baldridge2024imagen} enable high-fidelity text-to-image generation with strong prompt adherence.  Despite these advances, no prior work in aerial-view detection has leveraged such foundation models to construct controlled diagnostic benchmarks isolating failure modes. We introduce a framework that systematically employs \textit{Imagen~3}'s generative capabilities for interpretable diagnostic analysis and targeted data supplementation for aerial detectors.

\subsection{Benchmarking Limitations in Aerial Vision}

Object detection in aerial and satellite imagery remains challenging due to large-scale variation, arbitrary object orientations, and complex backgrounds~\cite{DOTAv2}. Consequently, even modern detectors (\eg, \textit{YOLOv8}~\cite{ultralytics2023yolov8}, \textit{DINO}~\cite{zhang2022dinodetrimproveddenoising}, and \textit{ViTDet}~\cite{li2022exploring}) still exhibit performance limitations on aerial benchmarks such as \textit{DOTAv2}~\cite{DOTAv2}, and their accuracy degrades under geographic domain shift~\cite{fang2025adapting, chen2018domain}. However, existing benchmarks cannot systematically diagnose these failures: \textit{DOTAv2} suffers from geographic and temporal concentration biases with strong statistical confounds; \textit{xView3}~\cite{paolo2022xview3} provides SAR imagery but lacks correlated RGB data; and \textit{RarePlanes}~\cite{shermeyer2021rareplanes} includes synthetic imagery via AI.Reverie's platform but introduces a simulation-to-real domain gap.

Domain adaptation methods have attempted to mitigate cross-domain discrepancies -- Fang~\etal~\cite{fang2025adapting} addresses geographic shifts via weak supervision, \textit{DA Faster}~\cite{chen2018domain} leverages domain discriminators, \textit{SWDA}~\cite{Saito_2019_CVPR} employs strong--weak alignment, and \textit{UMT}~\cite{deng2021unbiased} adopts a mean-teacher strategy---but these treat domain shift as a monolithic phenomenon without decomposing it into underlying causal factors. This indicates that adaptation alone is insufficient; advancing robust aerial detection requires a diagnostic perspective that disentangles and quantifies the impact of individual causal factors.

\section{Method}
\label{sec:method}

This section presents our methodology for diagnosing the weaknesses of pre-trained object detectors using synthetic aerial-view datasets generated by foundation generative models (\cref{sec:method:data_generation}). The process begins with the definition of an attribute taxonomy that guides data generation and ensures semantic coverage and visual diversity (\cref{sec:method:data_generation:taxonomy}). Synthetic data creation proceeds in two stages: (1)~initial image generation (\cref{sec:method:data_generation:image_generation}) and (2)~dataset enrichment through image editing (\cref{sec:method:data_generation:image_editing}). The resulting dataset is then annotated and used for model diagnosis based on attribute-conditioned evaluation (\cref{sec:method:model_diagnosis}).

\subsection{Diagnostic Dataset Generation}
\label{sec:method:data_generation}

The diagnostic dataset generation is a two-stage process. First, an initial set of synthetic images $\mathcal{I}_\text{G}$ is produced using a \textit{text-to-image} approach. To enhance diversity and achieve a more balanced distribution of visual attributes, a second stage performs \textit{image-and-text-to-image} generation, in which the initial images are edited based on textual prompts. This yields a complementary set of edited images $\mathcal{I}_\text{E}$. The final diagnostic dataset,
$\mathcal{I} = \mathcal{I}_\text{G} \cup \mathcal{I}_\text{E},$
is fully synthetic and annotated, providing improved coverage of attribute variations across scenes.
It is important to note that the presented generation framework is fundamentally model-agnostic; thus, open- or closed-source models can be incorporated, provided they produce data of sufficient quality and realism.

\subsubsection{Attribute Taxonomy}
\label{sec:method:data_generation:taxonomy}

\begin{figure}[t]
    \centering
    \includegraphics[width=\textwidth]{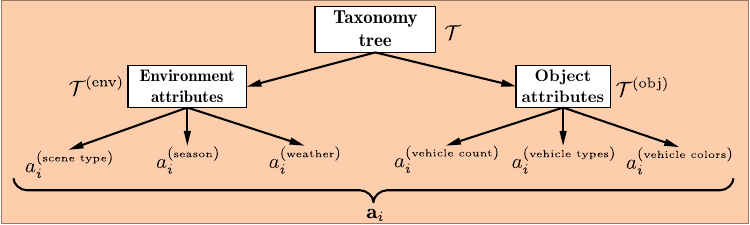}
    \caption{
        Attribute taxonomy tree and associated variables $a_i^{(c)}$ that constitute the attribute vector $\mathbf{a}_i$.
    }
    \label{fig:taxonomy}
\end{figure}

The attribute taxonomy $\mathcal{T}$ (\cref{fig:taxonomy}) defines the semantic structure underlying data generation. We deliberately designed this taxonomy to encompass the most critical axes of domain shift and dataset bias known to degrade aerial vision systems. It comprises three primary (environmental) attributes---\textit{scene type}, \textit{season}, and \textit{weather}---and three secondary (object-level) attributes---\textit{vehicle count}, \textit{vehicle color}, and \textit{vehicle type}.

The primary attributes globally determine the image content, enabling testing of macro-level vulnerabilities such as changes in background texture, illumination, and seasonal clutter that are severely imbalanced in standard benchmarks (\eg, the overrepresentation of sunny, urban scenes in DOTAv2). 
The purpose of the secondary attributes is to ensure object diversity.
Potentially, they could also enable fine-grained analysis of local detection challenges and testing the detector's limits with respect to foreground-background contrast, scale variance, and dense object crowding. 
All attributes are integrated into the diagnostic pipeline described in \cref{sec:method:model_diagnosis}. While we use this specific taxonomy to validate our framework and address known aerial detection bottlenecks, the proposed framework is generic and can be adapted to any taxonomy suitable for a given application.

\subsubsection{Generating the Initial Image Set}
\label{sec:method:data_generation:image_generation}

\begin{figure}[t]
    \centering
    \includegraphics[width=1\linewidth]{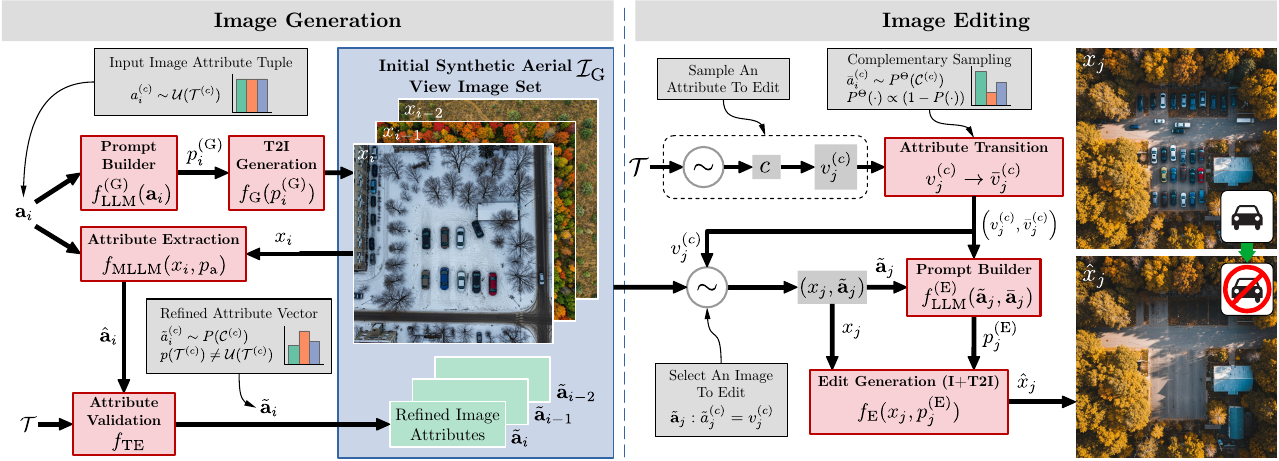}
    \caption{
        Overview of the synthetic data generation pipeline. 
        \textbf{Left:} Initial image generation using a \textit{text-to-image} foundation model $f_\text{G}$ conditioned on prompts $p_i^{(\text{G})}$ derived from uniformly sampled attribute values in the taxonomy $\mathcal{T}$. 
        \textbf{Right:} Image editing via \textit{image-and-text-to-image} generation to correct distributional biases and enrich underrepresented attribute values.
    }
    \label{fig:method_diagram}
\end{figure}

\Cref{fig:method_diagram}\textit{-left} illustrates the procedure for generating the initial image set. 
First, $N$ attribute tuples $\mathbf{a}_i = \{a_i^{(c)}\}_{c \in \mathcal{T}}$ are uniformly sampled from the taxonomy $\mathcal{T}$. 
Each tuple is provided to a \textit{Large Language Model} (LLM) $f_\text{LLM}$, which composes an aerial-view generation prompt $p_i^{(\text{G})}$ embedding the selected attribute values. 
The prompt is then passed to a \textit{text-guided diffusion model} (TDM) $f_\text{G}$ to generate the image $x_i$.

TDMs often deviate from the intended prompt $p_i^{(\text{G})}$ due to \textit{generation defects}~\cite{liu2024discovering, wang2024compositional}. 
To assess these deviations, we employ a \textit{Multimodal Large Language Model} (MLLM) $f_{\text{MLLM}}$ to analyze each image and infer its attribute vector $\tilde{\mathbf{a}}_i$. 
Because the predicted attributes may not align perfectly with the discrete categories in $\mathcal{T}$, an \textit{attribute validation} procedure is applied to refine them (\cref{alg:attrib_refining}). Namely, the text embedding $\tilde{e}_i^{(\text{c})}$ of the element $\tilde{a}_i^{(\text{c})}$ in $\tilde{\mathbf{a}}_i$ is compared with the text embeddings of all values $\mathcal{C}^{(\text{c})}$ in $\mathcal{T}$ corresponding to category $c$. The embeddings are predicted by a text encoder $f_\text{TE}$, and cosine similarity is used to perform the comparison. If an available value from $\mathcal{T}$ has cosine similarity with $\tilde{e}_i^{(\text{c})}$ higher than a threshold $\tau$,  $\tilde{a}_i^{(\text{c})}$ is swapped with that value. If no similarity higher than $\tau$ is found, $\tilde{a}_i^{(\text{c})}$ is added to $\mathcal{T}$ as a new category value.
This process ensures semantic consistency between visual content and attribute labels.

However, generation defects introduce imbalances in attribute distributions $P(\mathcal{C}^{(c)})$, despite the uniform sampling of input attributes for image generation. The image editing stage, described next, addresses these biases.

\subsubsection{Data Enrichment Through Image Editing}
\label{sec:method:data_generation:image_editing}

The image editing stage employs an \textit{image-and-text-to-image} process to mitigate attribute imbalance within $\mathcal{I}_\text{G}$ (\cref{fig:method_diagram}\textit{-right}). 
For each attribute category $c$, a source value $v_j^{(c)}$ and a target value $\bar{v}_j^{(c)}$ are selected, where the latter is sampled from the complementary distribution $P^{\Theta}(\mathcal{C}^{(c)})$, emphasizing underrepresented attribute values. 
An image $x_j$ is chosen from $\mathcal{I}_\text{G}$ with attribute value $v_j^{(c)}$, and an editing prompt $p_j^{(\text{E})}$ is generated by an LLM $f_\text{LLM}^{(\text{E})}$, which explicitly requires the model to preserve unchanged attributes while modifying only the selected one. For more details on the prompts used, refer to the supplementary material. 
The resulting edited image is analyzed by $f_{\text{MLLM}}$ to validate attributes, as in \cref{alg:attrib_refining}. 
Repeating this process $M$ times yields the enriched image set $\mathcal{I}_\text{E}$.

\begin{algorithm}[t]
    \caption{Synthetic Image Attribute Validation}
    \centering
    \begin{minipage}{0.85\linewidth}
        \begin{algorithmic}[1]
            \label{alg:attrib_refining}
            \scriptsize
            \Require Taxonomy tree $\mathcal{T}$, Models: $f_{\text{MLLM}}$, $f_{\text{TE}}$, image $x_i$, similarity threshold $\tau$
            \Ensure Refined attributes $\hat{\mathbf{a}}_i$
            \State $\mathcal{C}^{(\text{env})} \gets \textsc{ChildrenNames}(\mathcal{T}^{(\text{env})})$
            \State $\hat{\mathbf{a}}_i \gets \{\}$
            \For{$c$ in $\mathcal{C}^{(\text{env})}$}
                \State $\tilde{a}^{(c)}_i \gets \textsc{ExtractAttributeValue}(f_{\text{MLLM}}, x_i, c)$
                \State $\mathcal{C}^{(c)} \gets \textsc{ChildrenNames}(\mathcal{T}, c)$
                \State $\mathcal{E}^{(c)} \gets \textsc{ExtractTextEmbeddings}(f_{\text{TE}}, \mathcal{C}^{(c)})$
                \State $\tilde{\mathbf{e}}_i^{(c)} \gets \textsc{ExtractTextEmbeddings}(f_{\text{TE}}, \tilde{a}_i^{(c)})$
                \State $\mathcal{S}_i^{(c)} \gets \textsc{CosineSimilarities}(\mathcal{E}^{(c)}, \tilde{\mathbf{e}}_i^{(c)})$
                \State $k_\text{max} \gets \arg\max(\mathcal{S}_i^{(c)})$
                \If{$\mathcal{S}_i^{(c)}[k_\text{max}] \geq \tau$}
                    \State $\hat{a}^{(c)}_i \gets \mathcal{C}^{(c)}[k_\text{max}]$
                \Else
                    \State $\hat{a}^{(c)}_i \gets \tilde{a}^{(c)}_i$
                    \State $\mathcal{C}^{(c)} \gets \mathcal{C}^{(c)} \cup \tilde{a}^{(c)}_i$
                \EndIf
                \State $\hat{\mathbf{a}}_i \gets \hat{\mathbf{a}}_i \cup \hat{a}^{(c)}_i$
            \EndFor
            \State \Return $\hat{\mathbf{a}}_i$, updated $\mathcal{T}$
        \end{algorithmic}
    \end{minipage}
\end{algorithm}

\subsubsection{Annotations}
\label{sec:method:data_generation:annotations}

Generative models do not natively produce structured annotations such as bounding boxes. 
Pre-trained detectors are unsuitable for pseudo-labeling because they rely on real datasets and may inherit their inherent biases. 
Instead, foundation \textit{Vision–Language Models} (VLMs) and MLLMs are employed in a \textit{zero-shot detection} setting. 
Prompts instruct these models to output vehicle bounding box coordinates in structured formats (\eg, JSON objects).

Given the limited precision of zero-shot detection in remote sensing imagery~\cite{soni2025earthdial, fiaz2025geovlm}, predictions from multiple models are aggregated and refined through a \textit{human-in-the-loop} process. 
Human operators verify true-positive detections and discard false positives in a binary decision-making manner. In \cite{10.1007/978-3-319-10602-1_24}, the authors report that this approach takes, on average, about 1 second per bounding box, whereas drawing bounding boxes takes about 26 seconds per object \cite{266a7585fabc4f25a5c3994eb35b0ae8}. Thus, the semi-automatic strategy we employ yields a significant speedup in labeling while maintaining high annotation quality.

\subsection{Model Diagnosis}
\label{sec:method:model_diagnosis}

To analyze the behavior of a pre-trained detector, an attribute-driven evaluation strategy is employed. 
For each selected attribute in the taxonomy $\mathcal{T}$, the synthetic dataset $\mathcal{I}$ is partitioned into subsets according to attribute values.
The detector is then evaluated independently on each subset, enabling a fine-grained analysis of its sensitivity to specific visual factors.
\section{Experiments}
\label{sec:experiments}

\subsection{Evaluation Protocol}
\label{sec:experiments:protocol}

We evaluate the proposed method (\cref{sec:method}) using a protocol designed to measure both its diagnostic capability and the extent to which synthetic insights transfer to real data. As the main evaluation metric, we use \textit{Average Precision} (AP) with an \textit{IoU} threshold of 50\%, which is widely adopted in the field of object detection in aerial imagery \cite{10292941, drones9080549}.
The evaluation proceeds as follows:

\begin{enumerate}
    \item A pretrained aerial-view vehicle detector is evaluated on the synthetic diagnostic dataset, and both overall and attribute-conditioned APs are computed.
    \item Attributes with the lowest AP are identified as underperforming categories.
    \item Additional real training images corresponding to these categories are obtained from public sources, labeled, and incorporated into the training set, and the detector is retrained.
    \item The updated model is re-evaluated to assess whether the targeted data augmentation leads to the expected performance improvements.
\end{enumerate}

\subsection{Datasets}
\label{sec:experiments:implementation_details:datasets}

\begin{figure}[t]
    \centering
    \includegraphics[width=\linewidth]{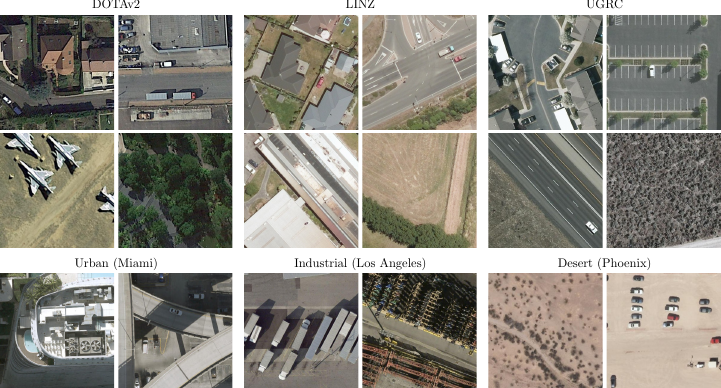}
    \caption{
        Samples from the real datasets used in our experiments
    }
    \label{fig:datasets}
\end{figure}

We used three real-world public multi-environment aerial/satellite top-down view image datasets for the initial vehicle detector pre-training: DOTAv2~\cite{DOTAv2}, LINZ~\cite{fang2025adapting}, and UGRC~\cite{fang2025adapting} (\cref{fig:datasets}).
As supplementary datasets, we collected and labeled three new real single-environment datasets: \textit{urban} (Miami, FL, USA), \textit{industrial} (Los Angeles, CA, USA), and \textit{desert} (Phoenix, AZ, USA) (\cref{fig:datasets}). 
All real datasets were sampled at $384\,\text{px}\times 384\,\text{px}$ with an effective \textit{ground sampling distance} (GSD) of $12.5\,\text{cm per px}$. 
Since \textit{Imagen 3} produces 1024~px images, they were scaled to 384~px to match the real data.
In total, 5,453 synthetic diagnostic images were generated, corresponding to 304 unique combinations of environmental attribute values.
More details are provided in \cref{tab:image_counts} and the supplementary material.

\begin{table} 
    \centering
    \scriptsize
    
    \renewcommand{\arraystretch}{1.15}
    \setlength{\tabcolsep}{6pt}
    \caption{Image count per dataset and split}
    \begin{tabular}{lrrrr}
        \toprule
        \multirow{2}{*}{\textbf{Dataset Name}} & \multicolumn{4}{c}{\textbf{Image count}} \\ 
        \cmidrule(lr){2-5}
         & \multicolumn{1}{c}{\textbf{Train}} & \multicolumn{1}{c}{\textbf{Val}} & \multicolumn{1}{c}{\textbf{Test}} & \multicolumn{1}{c}{\textbf{Total}} \\ 
        \midrule
        DOTAv2 & 28,458 & 8,807 & -- & 37,265 \\
        LINZ & 87,654 & 14,899 & 27,693 & 130,246 \\
        UGRC & 146,993 & 12,599 & 8,400 & 167,992 \\
        \midrule
        Urban (Miami) & 2,284 & -- & -- & 2,284 \\
        Industrial (LA) & 2,000 & -- & -- & 2,000 \\
        Desert (Phoenix) & 2,000 & -- & -- & 2,000 \\
        \midrule
        Imagen 3 & -- & -- & 5,453 & 5,453 \\
        \bottomrule
    \end{tabular}
    \label{tab:image_counts}
\end{table}

\subsection{Implementation Details}
\label{sec:experiments:implementation_details}

\begin{figure}
    \centering
    \includegraphics[width=1\linewidth]{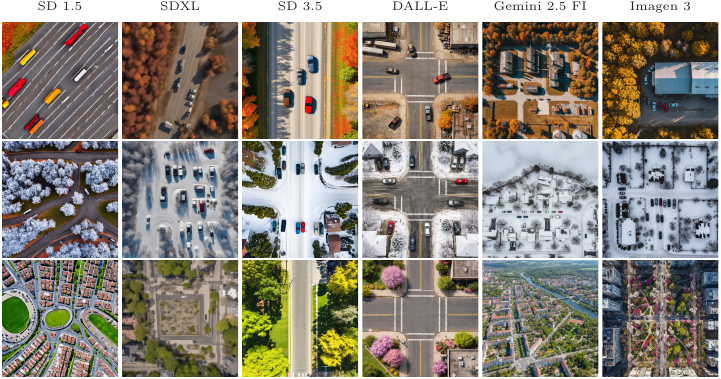}
    \caption{
        Qualitative comparison of six \textit{text-to-image} foundation generative models on the aerial-view image generation task.
        Each row is generated using the same prompt (see the supplementary material for more details). Open-source models: \textit{SD~1.5}~\cite{rombach2022high}, \textit{SDXL}~\cite{ICLR2024_081b0806}, and \textit{SD~3.5}~\cite{stabilityai_sd35}. Closed (commercial) models: \textit{DALL-E}~\cite{openai_dalle3}, \textit{Gemini 2.5 Flash Image}~\cite{comanici2025gemini}, \textit{Imagen~3}~\cite{baldridge2024imagen}. (\textit{Zoom in to see the details better.})
    }
    \label{fig:model_comparison}
\end{figure}

For synthetic data generation (\cref{sec:method:data_generation}), we employ large-scale foundation models accessed via cloud APIs, selected based on image quality and viewpoint consistency for nadir aerial imagery.
We evaluated six \textit{text-to-image} models for aerial scene synthesis (\cref{fig:model_comparison}). Open-source diffusion models, including \textit{Stable Diffusion 1.5}, \textit{Stable Diffusion XL}, and \textit{Stable Diffusion 3.5}~\cite{rombach2022high, ICLR2024_081b0806, stabilityai_sd35}, frequently produced unrealistic perspectives, inconsistent object scales, or weak prompt adherence when applied to aerial views without domain-specific fine-tuning. While fine-tuning may mitigate these issues, it reduces direct applicability.
Commercial models such as \textit{Gemini 2.5}\cite{comanici2025gemini} and \textit{Imagen 3}\cite{baldridge2024imagen} generated more consistent nadir-view scenes with stronger semantic fidelity (\cref{fig:model_comparison}). \textit{DALL-E}\cite{openai_dalle3} showed adequate visual quality but limited structural diversity across prompts. Prior evaluations\cite{10.5555/3600270.3602913} similarly report stronger photorealism and prompt alignment for \textit{Imagen}.
Based on these observations, we adopt \textit{Imagen 3}\cite{baldridge2024imagen} as the primary model for image generation and editing (\ie, $f_\text{G}$ and $f_\text{E}$), due to its consistent viewpoint control and photorealistic aerial synthesis.

Prompt composition for image generation ($f_\text{LLM}^{(\text{G})}$) and editing ($f_\text{LLM}^{(\text{E})}$) is performed with \textit{GPT-5}~\cite{openai2025gpt5} and \textit{Gemini 2.5 Flash Lite}~\cite{comanici2025gemini}, respectively. 
Image analysis and attribute extraction ($f_\text{MLLM}$) are carried out using \textit{Gemini 2.5 Flash}~\cite{comanici2025gemini}, while text embeddings for attribute matching ($f_\text{TE}$) are obtained via the \textit{Sentence-BERT} model~\cite{reimers2019sentence}. 
Vehicle bounding boxes are extracted using \textit{Gemini 2.5 Flash Lite} and \textit{Moondream 2}~\cite{vikhyatk2024moondream2}.

For the downstream evaluation, we consider three widely used aerial-view object detectors~\cite{10292941, drones9080549, DOTAv2}: \textit{Faster R-CNN (R50)}\cite{ren2015faster}, \textit{YOLOv8-M}\cite{ultralytics2023yolov8}, and \textit{ViTDet-B}\cite{li2022exploring}, representing single-stage, two-stage, and transformer-based detection paradigms, respectively. All experiments are implemented using \textit{MMDetection}\cite{mmdetection} and \textit{MMYOLO}~\cite{mmyolo2022}, and are conducted on a server equipped with eight NVIDIA Quadro RTX 6000 GPUs (24 GB VRAM each).

\subsection{Aerial-view Object Detectors Diagnosis}

\subsubsection{Data Annotations}

The first row of \cref{tab:bbox_approval_rates} reports the number of vehicle bounding boxes predicted by \textit{Moondream 2} and \textit{Gemini 2.5 Flash Lite} on our synthetic diagnostic dataset. We merge both prediction sets and average highly overlapping boxes to suppress duplicates, yielding a curated set of pseudo-annotations (reported under \textit{Total}). The last column presents the number and percentage of boxes approved by the human annotator. The same procedure is applied to the three supplementary real datasets, whose statistics are summarized in the bottom rows of \cref{tab:bbox_approval_rates}.
The approval rates underscore the need for human verification and indicate that current foundation models are insufficient for reliable zero-shot detection of aerial vehicles. Nevertheless, their ensemble provides an overcomplete proposal set that substantially accelerates annotation---by approximately 13$\times$, based on reported annotation speeds in \cite{10.1007/978-3-319-10602-1_24,266a7585fabc4f25a5c3994eb35b0ae8}.

\begin{table}[t]
    \centering
    \scriptsize
    \setlength{\tabcolsep}{3pt}
    \caption{Human operator bounding box approval rates}
    \begin{tabular}{lrrrr}
        \toprule
         \multicolumn{1}{c}{\textbf{Dataset Name}} & \multicolumn{1}{c}{\textbf{Moondream 2}} & \multicolumn{1}{c}{\textbf{Gemini 2.5 FL}} & \multicolumn{1}{c}{\textbf{Total}} & \multicolumn{1}{c}{\textbf{Human Approved}} \\ 
        \midrule
        Imagen 3 & 30,133 & 51,008 & 53,885 & 31,519 (58.5\%) \\
        \midrule
        Urban (Miami) & 7,606 & 13,597 & 14,974 & 9,741 (65.1\%) \\
        Industrial (LA) & 3,949 & 7,850 & 8,482 & 4,101 (48.3\%) \\
        Desert (Phoenix) & 499 & 1,312 & 1,429 & 982 (68.7\%) \\
        \bottomrule
    \end{tabular}
    \label{tab:bbox_approval_rates}
\end{table}

\subsubsection{Identifying Lowest-performing Scenes}

\begin{figure}
    \centering
    \includegraphics[width=1\linewidth]{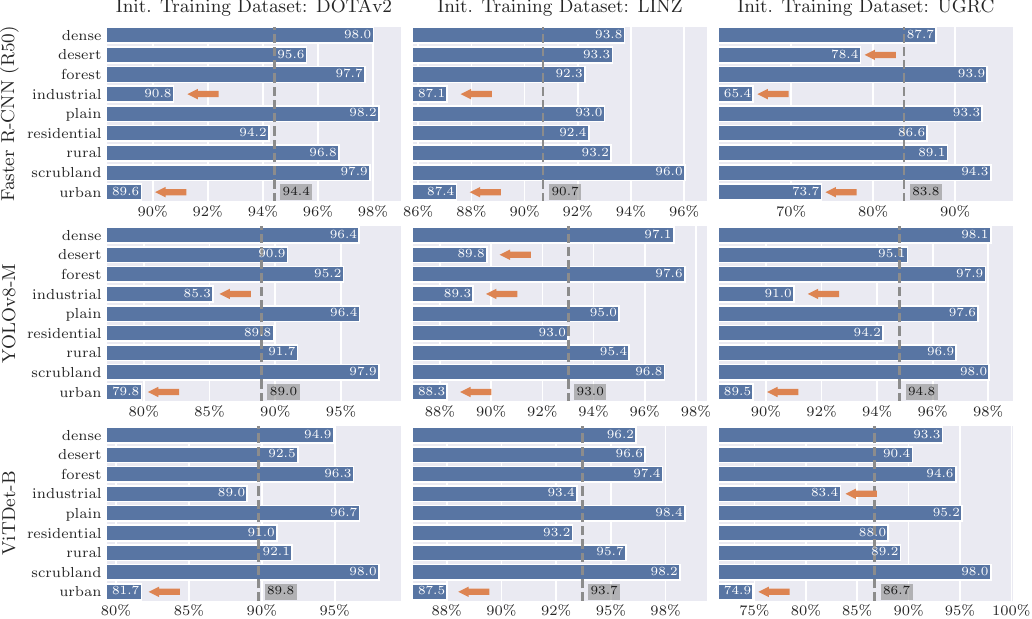}
    \caption{
        Scene-wise $\text{AP}^{50}$ of three detectors trained on real aerial datasets (DOTAv2, LINZ, UGRC) and evaluated on the synthetic diagnostic set. Rows denote models; columns denote training datasets. Orange arrows highlight scene types below the overall AP (gray dashed line). Best viewed in color.
    }
    \label{fig:model_diagnosis}
\end{figure}

To evaluate the diagnostic capability of our method, we trained three object detectors---\textit{Faster R-CNN}, \textit{YOLOv8}, and \textit{ViTDet}---on three real aerial-view datasets: \textit{DOTAv2}, \textit{LINZ}, and \textit{UGRC}, resulting in nine model–dataset combinations. Each detector was then evaluated on our synthetic diagnostic dataset to analyze performance across different \textit{scene types}, one of the key image attributes. 
The dataset includes nine scene categories: \textit{dense}, \textit{desert}, \textit{forest}, \textit{industrial}, \textit{plain}, \textit{residential}, \textit{rural}, \textit{scrubland}, and \textit{urban}. 
\Cref{fig:model_diagnosis} reports the scene-wise \textit{average precision} ($\text{AP}^{50}$) for all models. Orange arrows highlight the \textit{problematic} scene types where the detector performance drops by more than one percentage point relative to the dataset-level $\text{AP}^{50}$ (gray dashed line). 
The \textit{urban} scene has been marked as \textit{problematic} in all nine tests, while the \textit{industrial} scene in seven, and the \textit{desert} scene in two, respectively.

\subsubsection{Improvement Through Targeted Supplementation}

After identifying underperforming scene categories as described in the previous section, we conduct a targeted data supplementation experiment to validate the diagnostic findings. Specifically, we introduce three additional real-world datasets, each representing a single-scene environment (\ie, \textit{urban}, \textit{industrial}, and \textit{desert}), selected from the problematic scene set. These datasets are used to augment the initial training data (\textit{DOTAv2}, \textit{LINZ}, and \textit{UGRC}) and assess whether targeted real-data supplementation---guided by insights from our synthetic diagnostic analysis---leads to measurable improvements in detector performance.

\begin{table} 
\centering
\scriptsize
\setlength{\tabcolsep}{4.5pt}
\caption{
    Effect of supplementing training data with additional urban, industrial, and desert imagery.
}
\begin{tabular}{llcccccc}
\toprule
\multirow{2}{*}{
    \begin{tabular}[l]{@{}l@{}}
        \textbf{Initial}\\ 
        \textbf{Training}\\ 
        \textbf{Dataset}
    \end{tabular}} &
\multirow{2}{*}{\textbf{Model}} &
\multicolumn{3}{c}{\textbf{Supplementary Data}} &
\multicolumn{3}{c}{\textbf{Synthetic Data Testing (AP)}} \\
\cmidrule(lr){3-5} \cmidrule(lr){6-8}
 & & Urban & Industrial & Desert &
 Baseline & With suppl. & Gain \\
\midrule
\multirow{3}{*}{DOTAv2}
 & Faster R-CNN R50 & \cmark & \cmark &        & 93.8\% & 94.7\% & \cellcolor{green!20} +0.9\% \\
 & YOLOv8-M         & \cmark & \cmark &        & 87.2\% & 92.3\% & \cellcolor{green!20} +5.1\% \\
 & ViTDet-B         & \cmark &        &        & 86.3\% & 95.9\% & \cellcolor{green!20} +9.6\% \\
\midrule
\multirow{3}{*}{LINZ}
 & Faster R-CNN R50 & \cmark & \cmark &        & 87.7\% & 95.1\% & \cellcolor{green!20} +7.4\% \\
 & YOLOv8-M         & \cmark & \cmark & \cmark & 92.7\% & 95.8\% & \cellcolor{green!20} +3.1\% \\
 & ViTDet-B         & \cmark &        &        & 91.5\% & 96.1\% & \cellcolor{green!20} +4.6\% \\
\midrule
\multirow{3}{*}{UGRC}
 & Faster R-CNN R50 & \cmark & \cmark & \cmark & 80.9\% & 93.5\% & \cellcolor{green!20} +12.6\% \\
 & YOLOv8-M         & \cmark & \cmark &        & 94.2\% & 95.9\% & \cellcolor{green!20}  +1.7\% \\
 & ViTDet-B         & \cmark & \cmark &        & 81.6\% & 95.0\% & \cellcolor{green!20} +13.4\% \\
\bottomrule
\end{tabular}
\label{tab:supplement_ap}
\end{table}

\begin{figure}[t]
    \centering
    \includegraphics[width=1\linewidth]{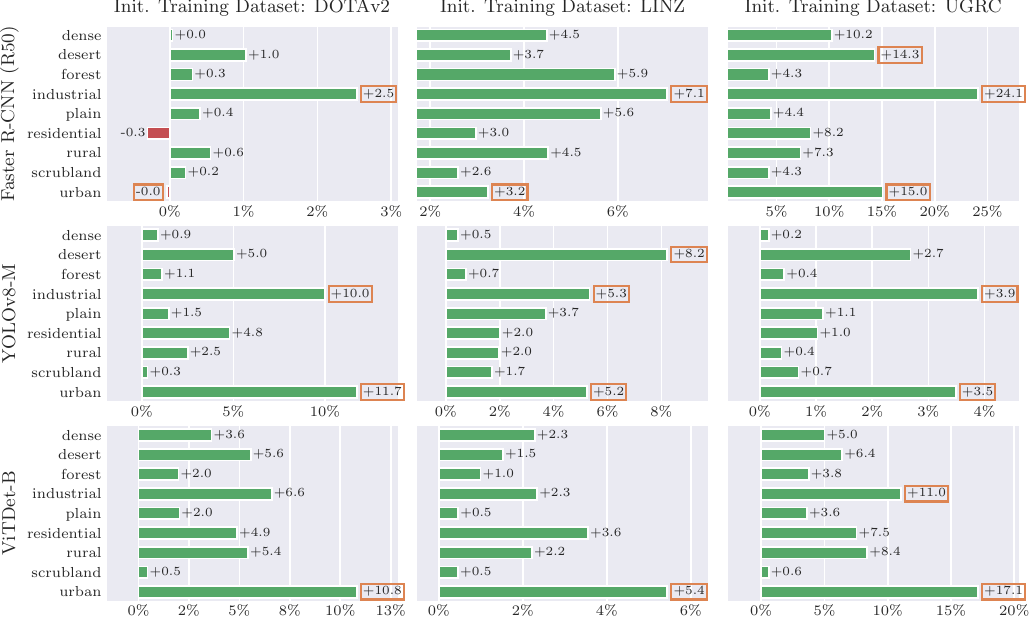}
    \caption{
        $\text{AP}^{50}$ gains per scene type provided by the data supplementation, described in \cref{tab:supplement_ap}. The values outlined in the orange box correspond to the scene types for which additional supplement data were used during supplementation. Best viewed in color.
    }
    \label{fig:model_supplementation_gains}
\end{figure}

\Cref{tab:supplement_ap} summarizes the impact of targeted supplementation across all nine model–dataset combinations. It is important to note that each supplementation dataset is at least an order of magnitude smaller than the training splits of the three initial datasets (\cref{tab:image_counts}).
In eight of the nine cases, augmenting the training data with real images from the identified weak scene types led to higher dataset-level $\text{AP}^{50}$, with improvements of up to 13\%. Only the \textit{Faster R-CNN} model trained on \textit{DOTAv2} showed no notable change.

\Cref{fig:model_supplementation_gains} provides a finer breakdown of scene-wise AP improvements. In most cases, the supplemented scene categories exhibit the largest relative performance gains, confirming that the insights derived from synthetic data effectively guide real-data collection.

\subsection{Targeted vs. Random Supplementation Experiments}

\begin{figure}[t]
    \centering
    \includegraphics[width=\linewidth]{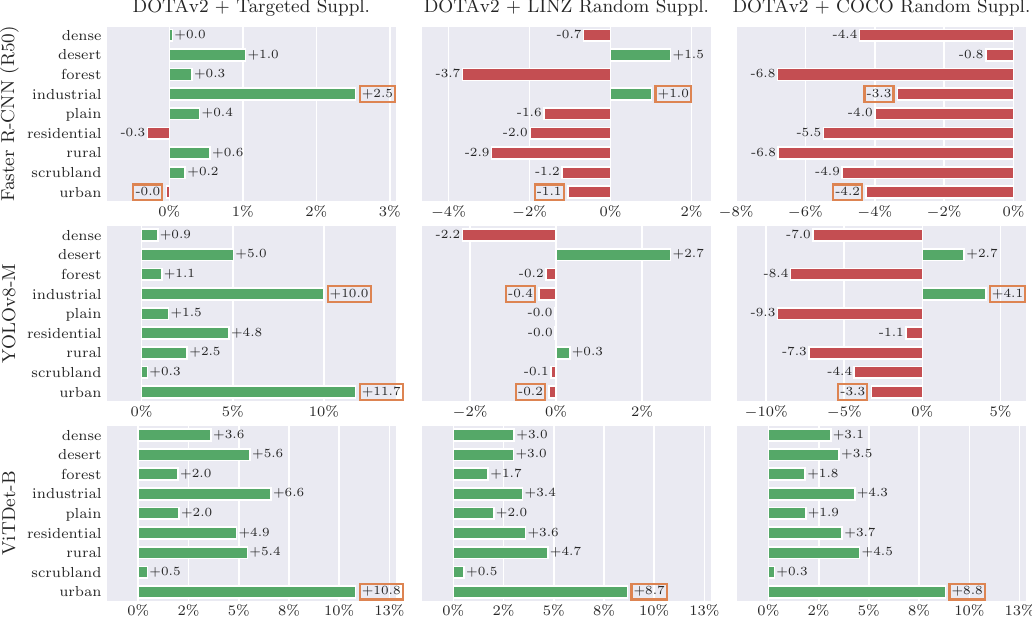}
    \caption{
        Targeted vs. random (non-targeted) supplementation AP gains
    }
    \label{fig:random_supplementation}
\end{figure}

\Cref{fig:random_supplementation} compares the AP gains achieved by targeted versus random (non-targeted) supplementation when \textit{DOTAv2} is used as the primary training dataset. To ensure a strictly controlled data budget, we randomly subsample two datasets---\textit{LINZ} (aerial) and \textit{COCO}~\cite{lin2014microsoft} (non-aerial)---to exactly match the image counts of our targeted supplementary datasets (\cref{tab:image_counts}).

As the results demonstrate, targeted supplementation guided by our diagnostic pipeline consistently outperforms random sampling. For standard architectures such as \textit{Faster R-CNN} and \textit{YOLOv8}, injecting randomly sampled data degrades performance across most scene types (as indicated by negative AP gains). This highlights the danger of blind data scaling, which often introduces domain interference and uninformative samples. In contrast, using the exact same data budget to target identified failure modes yields high-magnitude, positive improvements (e.g., up to +11.7\% for \textit{YOLOv8} in urban scenes). While the transformer-based \textit{ViTDet} exhibits greater baseline robustness to random data shifts, our targeted approach still achieves superior peak gains, proving the high data efficiency and practical value of the diagnostic loop.

\section{Conclusions}
\label{sec:conclusions}

This work demonstrated that foundation generative models can serve as diagnostic instruments for aerial-view object detection systems. By constructing a synthetic, attribute-controlled testbed, we identified systematic scene-dependent weaknesses of detectors trained on real datasets and showed that synthetic scene-wise trends closely reflect real-world performance gaps. Guided by these diagnostics, targeted supplementation with small, scene-specific real datasets improved performance by up to $13\%$ $\text{AP}^{50}$.

Although instantiated using commercial generative models, the proposed diagnostic loop is model-agnostic. Any generative and multimodal systems capable of producing high-quality, prompt-aligned aerial imagery can be integrated into the pipeline. We selected closed-source models for their current maturity and viewpoint consistency, without task-specific fine-tuning. However, reliance on proprietary systems raises reproducibility considerations, as such models may evolve or become unavailable. Open-source alternatives, while desirable for accessibility, often require substantial computational resources and domain adaptation to achieve comparable realism. As open-weight generative models mature, integrating them into this framework will further improve transparency and long-term reproducibility.

Our experiments focus on RGB nadir aerial imagery and vehicle detection. Extending synthetic diagnostic probing to additional viewpoints and sensing modalities—such as multispectral, thermal, or SAR imagery—remains an important direction. Similarly, while we defined a foundational taxonomy of scene attributes (scene type, season, weather) and performed diagnosis primarily along scene-type variations, broader, more granular studies of attributes are needed to assess cross-factor interactions and task generality.

More broadly, this study highlights controlled synthetic probing as a practical mechanism for improving remote sensing systems. Rather than replacing real data, synthetic diagnostics enable informed data acquisition by identifying where additional coverage is most impactful. In high-stakes Earth observation applications---where balanced data collection is costly, and deployment conditions are variable---such targeted guidance offers a principled path toward more reliable and resource-efficient aerial detection pipelines.
 

\section*{Acknowledgments}
This work has been funded by the DEVCOM Army Research Lab, USA. 

%
%
\bibliographystyle{splncs04}
\bibliography{main}

\clearpage
\appendix
\renewcommand{\thesection}{\Alph{section}}
\setcounter{page}{1}
\setcounter{section}{0}

\begin{center}
    {\Large \bfseries Diagnosing Aerial-View Object Detectors with Foundational Image Generative Models}\\[0.5em]
    {\large Supplementary Material}
\end{center}

\section{Real Data - Additional Information}

\subsection{Supplementary Real Data Sources}

\begin{figure}
  \centering
  \begin{subfigure}{0.49\linewidth}
    \centering
    \includegraphics[width=\linewidth]{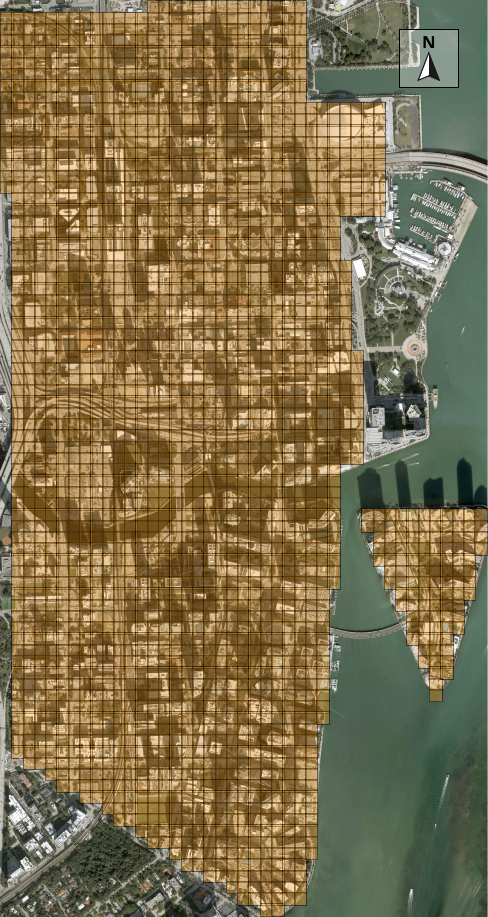}
    \caption{\textbf{Miami-Urban supplementary dataset}, sampled from Downtown Miami, FL, USA. The yellow squares depict the 2,284 image tiles, each of size $48\,\text{m} \times 48\,\text{m}$, with a 10\% overlap.}
    \label{fig:miami_urban_dataset}
  \end{subfigure}
  \hfill
  \begin{subfigure}{0.49\linewidth}
  \includegraphics[width=\linewidth]{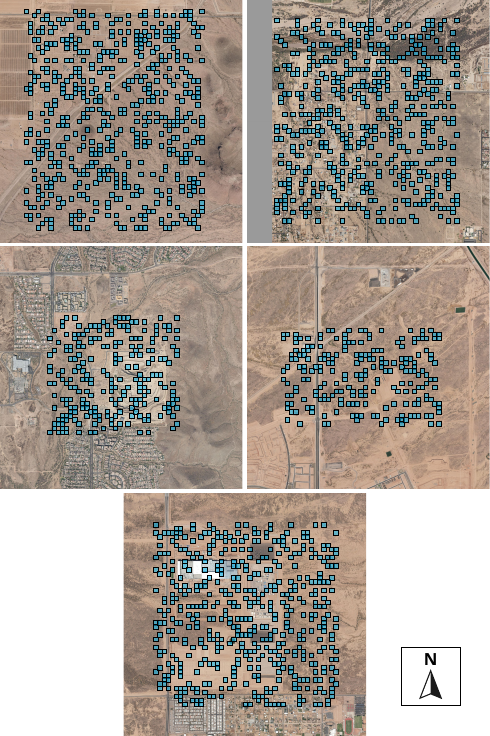}
    \caption{\textbf{Phoenix-Desert dataset} sampled from five peripheral locations around the City of Phoenix, Arizona, USA. The blue squares depict the exact sampling location of the 2,000 samples with size $48\,\text{m} \times 48\,\text{m}$.}
    \label{fig:phoenix_desert_dataset}
  \end{subfigure}
  \caption{Urban and desert real supplementary datasets}
  \label{fig:short}
\end{figure}

\begin{figure*}
    \centering
    \includegraphics[width=1\linewidth]{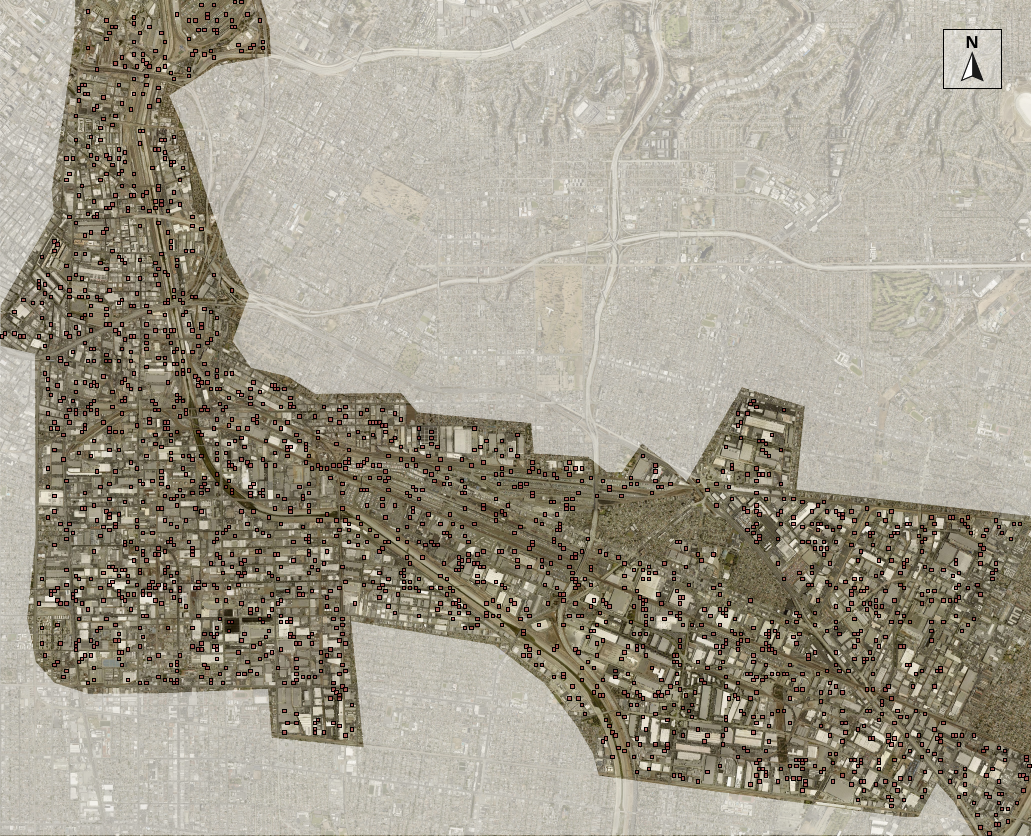}
    \caption{\textbf{Los Angeles - Industrial dataset} sampled from LARIAC6 source. The red squares depict the exact sampling locations of the 2,000 samples with size $48\,\text{m} \times 48\,\text{m}$.}
    \label{fig:la_industrial_dataset}
\end{figure*}

We utilized QGIS\footnote{\url{https://qgis.org/}} software to download the three real supplementary datasets. 
The \textit{Miami - Urban dataset} (\cref{fig:miami_urban_dataset}) has been sourced from the \textit{ESRI World Imagery}\footnote{\url{https://www.arcgis.com/home/item.html?id=10df2279f9684e4a9f6a7f08febac2a9}}.
The \textit{Los Angeles - Industrial} (\cref{fig:la_industrial_dataset}) dataset represents a part of \textit{LARIAC6-09 (2022 Summer Urban RGB Ortho)}\footnote{\url{https://www.arcgis.com/home/item.html?id=a1dc35b23d7849098f71e56841da8148}}.
The \textit{Phoenix - Desert} (\cref{fig:phoenix_desert_dataset}) dataset has been sampled from \textit{City of Phoenix 2024 Aerial Imagery}\footnote{\url{https://www.arcgis.com/home/item.html?id=0e7011b550e94ab99bc351a633d24952}}.
The figures also depict the exact sampling locations of the extracted image tiles.

\subsection{Annotations}

\begin{figure}
    \centering
    \includegraphics[width=0.5\linewidth]{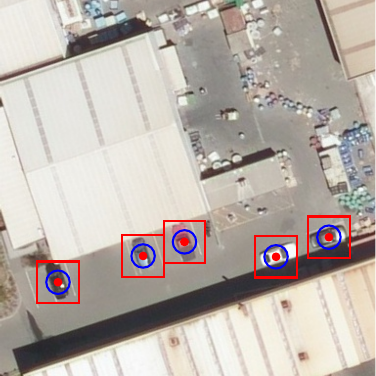}
    \caption{An example of a LINZ dataset image with square bounding boxes for center-based object detection. 
    The red dots depict the annotated vehicle centers, and the blue circles depict the 12 px decision circles.}
    \label{fig:bounding_boxes}
\end{figure}

The LINZ and UGRC datasets provide object location annotations ($x$ and $y$ coordinates of the object center), rather than bounding boxes. In order to employ bounding box-based object detectors, we center square bounding boxes at each object center with size $42.36\,\text{px} \times 42.36 \, \text{px}$ (\cref{fig:bounding_boxes}). Combined with \textit{IoU} threshold of 50\%, this size ensures that all positive prediction centers will fall within a circle with radius $12\, \text{px}$ centered at the annotations. Given the ground sampling distance of the data ($0.125$ m per px) and the average physical size of a small vehicle, such as a car or a pickup truck, $\approx 5 \, \text{m} \times \approx 2\, \text{m}$, the equivalent image-space size of the vehicle will be $\approx 40 \, \text{px} \times \approx 16\, \text{px}$. Thus, the 12 px decision circle delivers a practically sufficient level of spatial precision.
This motivates our use of $\text{AP}^{50}$ as the primary metric, as tighter IoU thresholds would penalize spatial noise irrelevant to the task.
To make all experiments equivalent, we replace the original bounding boxes from DOTAv2, Miami-Urban, LA-Industrial, and Phoenix-Desert with square bounding boxes following the same principle. 

\subsection{Scene Types Distribution}

\begin{figure*}
    \centering
    \begin{subfigure}{\textwidth}
        \includegraphics[width=1\linewidth]{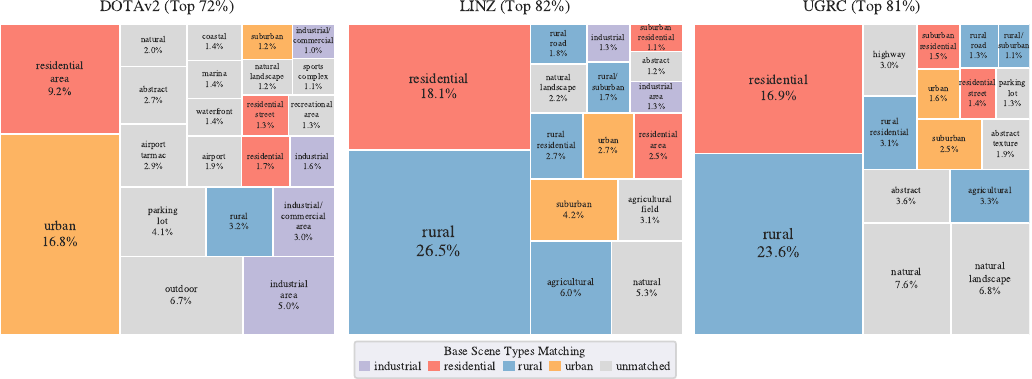}
        \caption{Raw scene types as predicted by \textit{Gemini 2.5 Flash} (treemaps).}
        \label{fig:scene_type_distr_real_raw}
    \end{subfigure}
    
    \vspace{1mm}
    
    \begin{subfigure}{\textwidth}
        \includegraphics[width=1\linewidth]{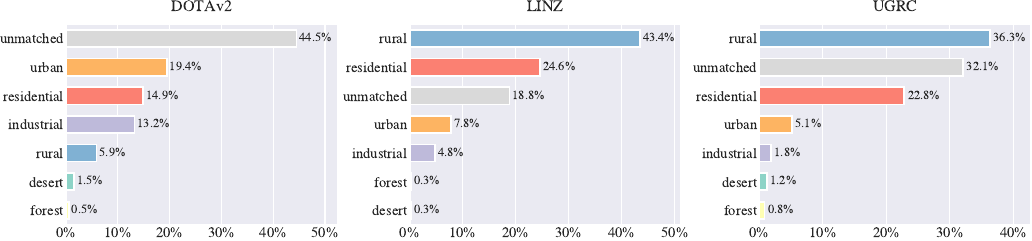}
        \caption{Scene types after cosine similarity matching and thresholding with the nine base categories.}
        \label{fig:scene_type_distr_real_matched}
    \end{subfigure}
    \caption{
        Scene types distributions of the three pre-training real aerial view datasets used in our experiments.
    }
    \label{fig:scene_type_distr_real}
\end{figure*}

\begin{figure*}
    \centering
    \begin{subfigure}{\textwidth}
        \includegraphics[width=1\linewidth]{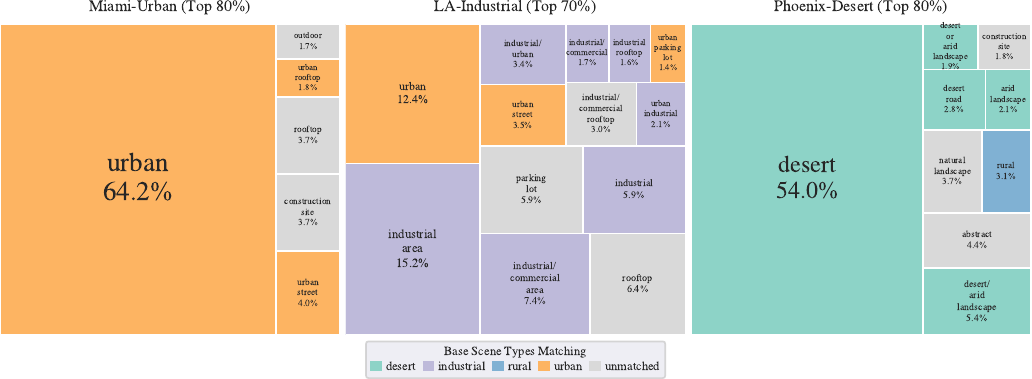}
        \caption{Raw scene types as predicted by \textit{Gemini 2.5 Flash} (treemaps)}
        \label{fig:scene_type_distr_real_supp_raw}
    \end{subfigure}
    
    \vspace{1mm}
    
    \begin{subfigure}{\textwidth}
        \includegraphics[width=1\linewidth]{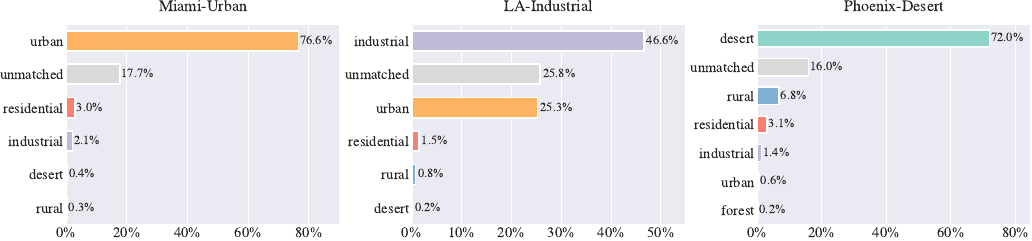}
        \caption{Scene types after matching with the nine base categories}
        \label{fig:scene_type_distr_real_supp_matched}
    \end{subfigure}
    \caption{
        Scene types distributions of the three full supplemental real aerial view datasets used in our experiments.
    }
    \label{fig:scene_type_distr_real_supp}
\end{figure*}

\cref{fig:scene_type_distr_real_raw} reports the distribution of the predicted raw\footnote{As predicted by the \textit{Gemini 2.5 Flash}. Cosine similarity-based matching to base categories is performed for colorization.} image scene types for each of the three pre-training real datasets (DOTAv2, LINZ, and UGRC). We analyze a uniformly sampled subset of each dataset to reduce computational cost while preserving representativeness.
Each plot represents the majority of the analyzed images, with the bottom 18\%-28\% excluded because they are hard to fit on the graphs.
From the figure, it is evident that the LINZ and UGRC datasets have similar distributions of scene types, with the top-2 categories — \textit{rural} and \textit{residential} — matching. Together they represent 44.6\% and 40.5\% of the datasets, respectively. On the other hand, \textit{urban} and \textit{residential area} are the two most popular scene types in DOTAv2, accounting for approximately 26\% of the data. 
The colorization of the blocks indicates the base taxonomy scene-type category to which they were matched, based on the cosine similarity of their text embeddings.
\cref{fig:scene_type_distr_real_matched} visualizes the histograms of the matched categories. We can see that 44.5\% of the samples in the DOTAv2 subset were not matched to any taxonomy scene type, whereas this percentage is lower in LINZ and UGRC. 

\cref{fig:scene_type_distr_real_supp} illustrates the distribution of the scene types in the supplementary datasets. We can clearly see that a single category (urban, industrial, or desert) dominates each.

\section{Synthetic Data - Additional Information}

\subsection{Complete Attribute Taxonomy}
\label{sec:complete_taxonomy}

\Cref{tab:taxonomy_full_env} and \Cref{tab:taxonomy_full_obj} provide complete information for the environmental and object taxonomy tree nodes (\cref{fig:taxonomy}) and their values. The initial version of the taxonomy was created with the help of \textit{ChatGPT}\footnote{\url{https://chatgpt.com/}} and subsequently refined by manually adding or removing categories deemed irrelevant. For instance, the initial weather-condition set included the \textit{cloudless} category, which we empirically estimated overlaps significantly with the \textit{sunny} category. Unlike the other attributes, which are single-valued, vehicle types and vehicle colors are multi-valued: a single image can have instances of different vehicle types, and each vehicle type can have different colors. 

\begin{table}
    \centering
    \scriptsize
    \setlength{\tabcolsep}{8pt}
    \renewcommand{\arraystretch}{1.15}
    \caption{Environmental attributes and values}
    \begin{tabular}{p{0.18\linewidth}p{0.2\linewidth}p{0.16\linewidth}p{0.18\linewidth}}
    \toprule
    \multicolumn{2}{c}{\textbf{Scene types}} &
    \multicolumn{1}{c}{\textbf{Seasons}} &
    \multicolumn{1}{c}{
        \begin{tabular}[c]{@{}c@{}}
            \textbf{Weather}\\[-1pt]\textbf{Conditions}
        \end{tabular}
    } \\
    \midrule
    \vspace{-2\baselineskip}
    \begin{itemize}[leftmargin=*]
        \item dense
        \item desert
        \item forest
        \item industrial
        \item plain
    \end{itemize}
    &
    \vspace{-2\baselineskip}
    \begin{itemize}[leftmargin=*]
        \item residential
        \item rural
        \item scrubland
        \item urban
    \end{itemize}
    &
    \vspace{-2\baselineskip}
    \begin{itemize}[leftmargin=*]
        \item spring
        \item summer
        \item fall
        \item winter 
    \end{itemize}
    &
    \vspace{-2\baselineskip}
    \begin{itemize}[leftmargin=*]
        \item cloudy
        \item dry
        \item snowy
        \item sunny
    \end{itemize}
    \\
    \bottomrule
    \end{tabular}
    \label{tab:taxonomy_full_env}
\end{table}

\begin{table} 
    \caption{Object attributes and values}
    \centering
    \scriptsize
    \setlength{\tabcolsep}{8pt}
    \renewcommand{\arraystretch}{1.15}
    \begin{tabular}{p{0.1\linewidth}p{0.2\linewidth}p{0.14\linewidth}p{0.12\linewidth}}
    \toprule
    \multicolumn{1}{c}{\textbf{Vehicle Count}} &
    \multicolumn{1}{c}{\textbf{Vehicle Types}} &
    \multicolumn{2}{c}{
        \textbf{Vehicle Colors}
    } \\
    \midrule
    \multicolumn{1}{c}{0-15} 
    &
    \vspace{-2\baselineskip}
    \begin{itemize}[leftmargin=*]
        \item car
        \item pickup truck
    \end{itemize}
    &
    \vspace{-2\baselineskip}
    \begin{itemize}[leftmargin=*]
        \item blue
        \item red
        \item brown
        \item gray
    \end{itemize} 
    &
    \vspace{-2\baselineskip}
    \begin{itemize}[leftmargin=*]
        \item green
        \item black
        \item white
        \item camo
    \end{itemize} 
    \\
    \bottomrule
    \end{tabular}
    \label{tab:taxonomy_full_obj}
\end{table}

\subsection{Image Generation}

\subsubsection{Prompt composition}

To produce the list of final image generation prompts, we use a two-stage approach. Initially, we uniformly and randomly sample $N$ attribute combinations for each taxonomy category, using the values in \Cref{tab:taxonomy_full_env} and \Cref{tab:taxonomy_full_obj}. Thus, each generated image will have a scene type, season, weather, vehicle count, a list of vehicle types, and a list of vehicle colors. 
The second step is to query OpenAI's GPT-5 model by feeding the set of attributes for each image and providing general instructions, which look like this: 

\begin{tcolorbox}
    \scriptsize
    
    You are a prompt engineer generating strictly realistic aerial top-down (nadir, ~90° overhead) image prompts.

    Rules for EVERY row:
    \begin{itemize}
        \item View: nadir / straight-down; no tilt or oblique.
        \item Style: photorealistic; not cinematic or artistic; avoid camera/lens jargon and brand names.
        \item Content: plausible overhead scale; vehicles can be small (parked or moving).
        \item Lighting: neutral daylight unless weather/season strongly implies otherwise.
        \item Consistency: scene type, weather, and season must be coherent.
        \item Vehicle colors: if provided as a list, reflect them; if missing, choose common realistic colors.
        \item Output: 1–2 sentences for "prompt". Optionally include a terse "negative\_prompt" (\eg, "no tilt, no fisheye, no motion blur, no text").
    \end{itemize}
    Return only the structured object that matches the schema I provided.
\end{tcolorbox}

Below are three examples of final image generation prompts produced with the method described above:

\begin{itemize}
    \item \textbf{Example 1:} Scene type: "rural", Season: "winter", Weather: "cloudy", Vehicle count: 2, Vehicle colors: [red, white]\\
    \textit{"Straight-down aerial view of a rural homestead in winter under overcast light, dormant fields with patchy snow, a farmhouse with a large white propane tank beside it; two small vehicles in red and white parked on the gravel drive."}
    
    \item \textbf{Example 2:} Scene type: "residential", Season: "spring", Weather: "sunny", Vehicle count: 9, Vehicle colors: [blue, brown, camo, gray, red, white]\\
    \textit{"High‑resolution aerial top‑down photograph of a spring residential landscape under sunny conditions, featuring nine small vehicles (blue, brown, camo, gray, red, white) such as cars or pickup trucks, strictly realistic appearance, true color, directly overhead perspective."}
    
    \item \textbf{Example 3:} Scene type: "industrial", Season: "spring", Weather: "sunny", Vehicle count: 0, Vehicle colors: []\\
    \textit{"High‑resolution aerial top‑down photograph of a spring industrial landscape under sunny conditions, no visible vehicles, strictly realistic appearance, true color, directly overhead perspective."}
\end{itemize}

\subsubsection{Models Comparison Prompts}

\Cref{fig:model_comparison} from the main paper presents a qualitative comparison between six image generative models. Below is the list of the exact prompts used to generate those images with each model:

\begin{itemize}
    \item \textbf{Prompt 1:} \textit{"High‑resolution aerial top‑down photograph of a fall industrial landscape under sunny conditions, featuring five small vehicles (brown, gray, red) such as cars or pickup trucks, strictly realistic appearance, true color, directly overhead perspective."}

    \item \textbf{Prompt 2:} \textit{"High‑resolution aerial top‑down photograph of a winter residential landscape under cloudy conditions, featuring 11 small vehicles (black, blue, brown, gray, red, white) such as cars or pickup trucks, strictly realistic appearance, true color, directly overhead perspective."}

    \item \textbf{Prompt 3:} \textit{"High‑resolution aerial top‑down photograph of a spring urban landscape under sunny conditions, no visible vehicles, strictly realistic appearance, true color, directly overhead perspective."}
\end{itemize}

\subsection{Image Editing}

\begin{figure*}
    \centering
    \includegraphics[width=1\textwidth]{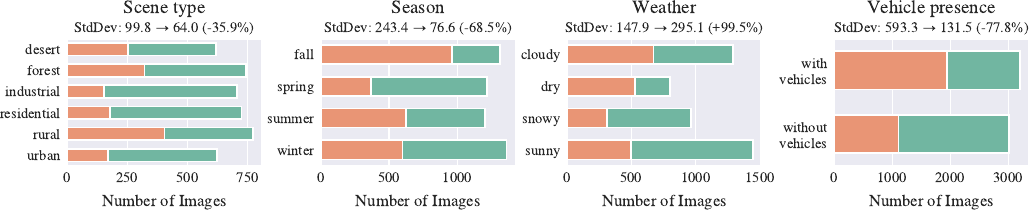}
    \caption{\textbf{Image attribute distributions} before \textit{(orange)} and after \textit{(green)} the image editing. The histogram bin standard deviations \textit{(before editing $\rightarrow$ after editing)} are depicted under each plot's title.
    }
    \label{fig:image_editing_hists}
\end{figure*}

\cref{fig:image_editing_hists} contains histograms illustrating the effects of image editing on the distribution of attribute values across the generated images. We use the \textit{standard deviation} across the bins of each histogram to assess the level of non-uniformity. From the provided values, it is evident that the complementary attribute sampling during the editing process, described in \cref{sec:method:data_generation:image_editing} of the main paper, had a positive effect on \textit{scene type}, \textit{season}, and \textit{vehicle presence} attribute values. They recorded a drop of 35.9\% to 77.8\%. On the contrary, the standard deviation of the \textit{weather} attribute almost doubled. We speculate that this is due to the generative model's inability to produce certain attribute combinations, likely due to inherent biases.

\begin{figure*}
    \centering
    \includegraphics[width=1\textwidth]{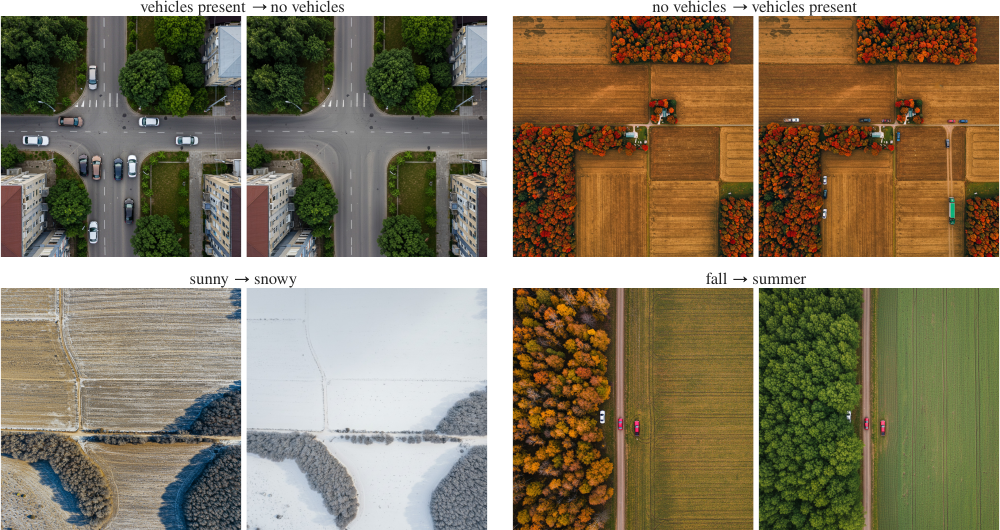}
    \caption{\textbf{Image editing examples:} removing or adding vehicles \textit{(first row)}, changing the weather conditions \textit{(bottom left)}, and changing the season \textit{(bottom right)}.
    }
    \label{fig:image_editing_examples}
\end{figure*}

\cref{fig:image_editing_examples} illustrates four image-editing examples that successfully remove and add vehicles to the scene, change the weather conditions (from a sunny winter scene to a cloudy and snowy winter scene), and change the season (from fall to summer).

\begin{figure*}
    \centering
    \includegraphics[width=1\textwidth]{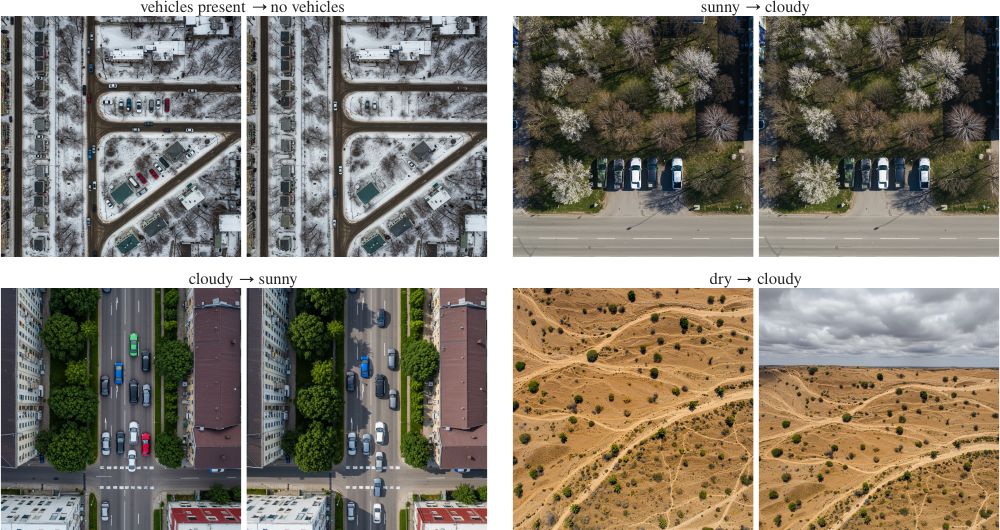}
    \caption{\textbf{Image editing defects:} failure to remove all vehicles \textit{(top left)}, failure to change the weather from sunny to cloudy \textit{(top right)}, failure to preserve the vehicle locations and colors while performing weather editing \textit{(bottom left)}, mixing two different camera views \textit{(bottom right)}.
    }
    \label{fig:image_editing_failure_examples}
\end{figure*}

\cref{fig:image_editing_failure_examples} illustrates four image-editing failures. In some cases, the image editing model failed to remove all vehicles from the scene. Our observations indicate that this behavior is more common when the cars are small and the model struggles to achieve fine-grained control. 
Sometimes the model does not perform any noticeable edits, such as failing to change the weather conditions from sunny to cloudy.
Another mistake the model makes is failing to limit the edits to the requested attributes. For instance, it failed to preserve the original cars shown in the image when changing the weather from cloudy to sunny (bottom-left example in \cref{fig:image_editing_failure_examples}). Lastly, sometimes the generated images contain an inconsistent mixture of two distinct camera views---vertical top-down and horizontal.

\end{document}